\documentclass[10pt,twocolumn,letterpaper]{article}

\usepackage[pagenumbers]{cvpr} 
\usepackage{amsmath}
\usepackage{soul}
\usepackage{booktabs}
\usepackage{overpic}
\usepackage{amssymb}
\usepackage{stfloats}
\usepackage[accsupp]{axessibility}
\usepackage{pifont}
\usepackage{multirow}
\usepackage{booktabs}
\usepackage{xcolor}         
\usepackage{colortbl}
\definecolor{mydarkred}{rgb}{0.6,0,0}
\definecolor{mydarkgreen}{rgb}{0,0.6,0}
\usepackage{threeparttable}
\usepackage{bm,bbm}
\usepackage{amsfonts}     
\usepackage{xcolor}          
\usepackage{tabularray}      
\UseTblrLibrary{booktabs}
\usepackage{algorithm}
\usepackage{algpseudocode}
\usepackage{tabularx}

\usepackage[dvipsnames]{xcolor}

\definecolor{cvprblue}{rgb}{0.21,0.49,0.74}
\usepackage[pagebackref,breaklinks,colorlinks,citecolor=cvprblue]{hyperref}

\def\paperID{15566} 
\def\confName{CVPR}
\def\confYear{2026}

\title{Late-decoupled 3D Hierarchical Semantic Segmentation with Semantic Prototype Discrimination based Bi-branch Supervision}

\author{
Shuyu Cao$^{1,\ddagger}$,
Chongshou Li$^{1,\ddagger}$,
Jie Xu$^{2,*}$,
Tianrui Li$^1$,
Na Zhao$^2$
\\
{$^1$School of Computing and AI, Southwest Jiaotong University, Chengdu, China} \\
{$^2$Singapore University of Technology and Design, Singapore}
}

\begin{document}

\maketitle

\let\thefootnote\relax\footnotetext{$^{\ddagger}$Equal contribution; $^{*}$Corresponding Author.}

\begin{abstract}
3D hierarchical semantic segmentation (3DHS) is crucial for embodied intelligence applications that demand a multi-grained and multi-hierarchy understanding of 3D scenes.
Despite the progress, previous 3DHS methods have overlooked following two challenges:
I) multi-label learning with a parameter-sharing model can lead to multi-hierarchy conflicts in cross-hierarchy optimization, and
II) the class imbalance issue is inevitable across multiple hierarchies of 3D scenes, which makes the model performance become dominated by major classes.
To address these issues, we propose a novel framework with a primary 3DHS branch and an auxiliary discrimination branch.
Specifically, to alleviate the multi-hierarchy conflicts, we propose a late-decoupled 3DHS framework which employs multiple decoders with the coarse-to-fine hierarchical guidance and consistency.
The late-decoupled architecture can mitigate the underfitting and overfitting conflicts among multiple hierarchies and can also constrain the class imbalance problem in each individual hierarchy.
Moreover, we introduce a 3DHS-oriented semantic prototype based bi-branch supervision mechanism, which additionally learns class-wise discriminative point cloud features and performs mutual supervision between the auxiliary and 3DHS branches, to enhance the class-imbalance segmentation.
Extensive experiments on multiple datasets and backbones demonstrate that our approach achieves state-of-the-art 3DHS performance, and its core components can also be used as a plug-and-play enhancement to improve previous methods.
\end{abstract}

\section{Introduction}\label{sec:intro}
3D semantic segmentation (3DS) is a fundamental task in 3D vision, aiming to assign a single semantic label to every point in a 3D point cloud.
Pioneering 3DS methods leverage point cloud oriented backbone networks such as PointNet~\cite{qi2017pointnet}, PointNet++~\cite{qi2017pointnetplusplus}, Point Transformer~\cite{zhao2021pointtransformer, guo2021pct, wu2024ptv3} to extract  point cloud features and use a classifier to learn point-wise labels.
In practice, however, an object usually possesses rich semantic meanings that go beyond a single label, e.g., a table belongs to both the furniture category and the wood products category, making traditional single-hierarchy 3DS models unable to satisfy the requirements in real-world applications.
This detailed, hierarchy-aware scene understanding is crucial for a wide range of future applications, from enabling safer navigation in autonomous driving and robotics~\cite{behley2019semantickitti,milioto2019rangenetplusplus} to creating richer interactive experiences in augmented reality~\cite{dai2017scannet, mccormac2017semanticfusion}.
To this end, the field has evolved towards 3D hierarchical semantic segmentation (3DHS) aiming at learning multi-hierarchy semantic labels for every point, and 3DHS tasks have recently garnered significant attention from the research community~\cite{li2020campus3d,li2025deep_hierarchical_learning}.


Due to the design of a single hierarchy, previous 3DS models are architecturally ill-suited for the 3DHS task. Moreover, training an individual 3DS model for each hierarchy is costly and there is a lack of positive interaction among multiple hierarchies~\cite{yu2019partnet_recursive,roberts2021lsd_structurenet,mo2019structurenet}.
Therefore, researchers have devoted their efforts to innovating in the hierarchical model structure and cross-hierarchy interaction of 3DHS.
For example, pioneering work designed a multi-task hierarchical segmentation network (MTHS~\cite{li2020campus3d}) that stacks multiple parallel decoders and classifiers on a shared point cloud encoder and leverages a multi-task classification loss to achieve 3DHS.
Furthermore, \citet{li2025deep_hierarchical_learning} proposed a model-agnostic deep hierarchical learning (DHL) method, which adopts the top-bottom and bottom-top fusion mechanisms and optimizes the cross-hierarchy consistency loss to pass information across hierarchies.



Although existing 3DHS methods achieved important progress, they still face the following two challenges.
I)~\emph{Multi-hierarchy conflicts}:
In 3DHS, previous methods usually learn multiple labels for different hierarchies with the shared decoder parameters~\cite{qi2017pointnetplusplus,qi2017pointnet,zhao2021pointtransformer,li2025deep_hierarchical_learning}.
They often require manually designed balancing strategies for different hierarchies' label learning (e.g., distinct weights for hierarchical objectives).
However, there are conflicts among hierarchies, which makes balancing difficult and limits the model's effectiveness.
II)~\emph{Class imbalance issue}:
The point class imbalance is inevitable as the number of hierarchies increases~\cite{li2020campus3d}, resulting in the models being easily dominated by majority classes (e.g., wall, floor) and their segmentation performance on minority classes (e.g., column, board) degrades.


\begin{figure*}[!t]
\centering
\includegraphics[width=1\linewidth]{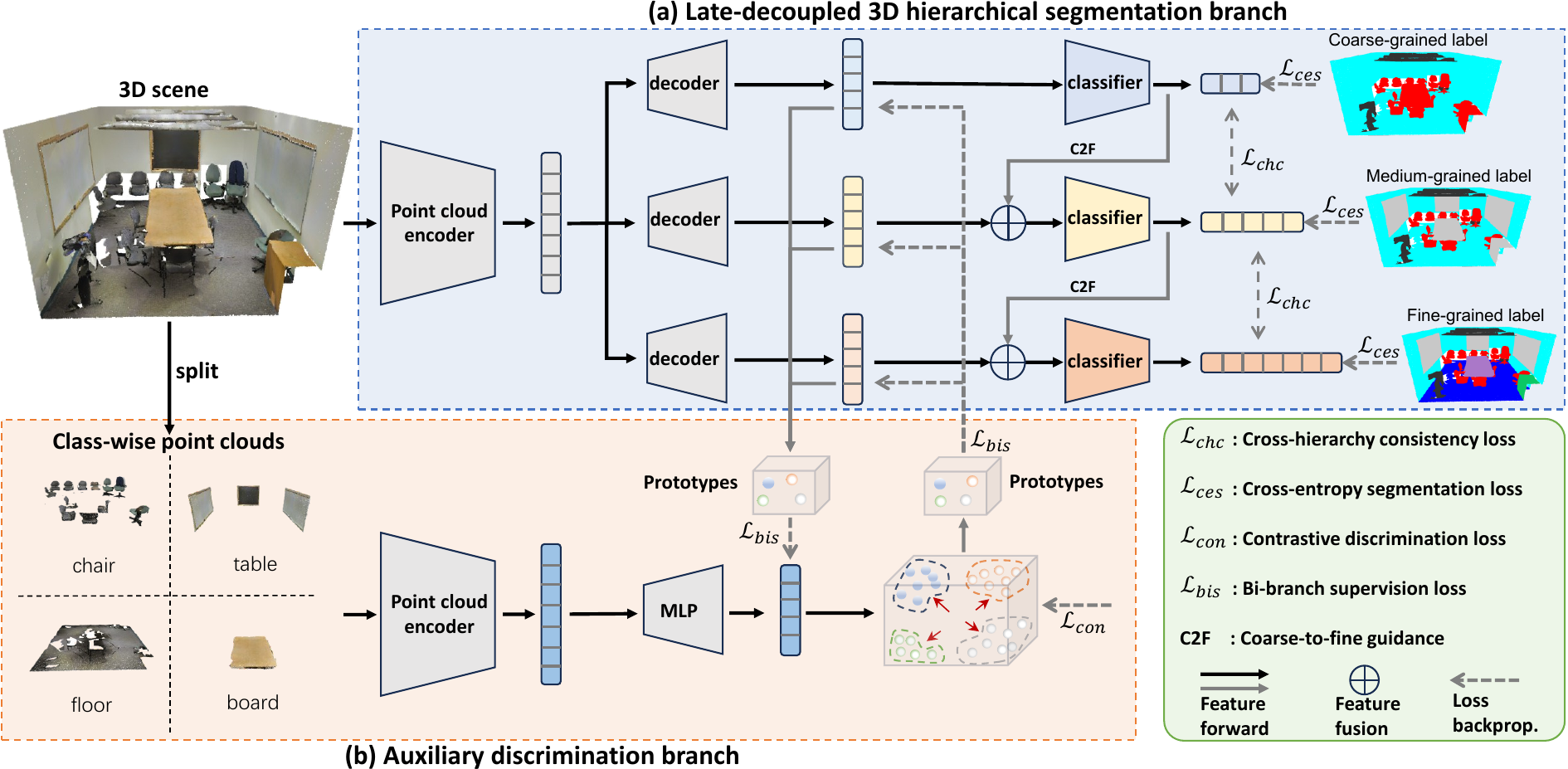}
\vspace{-0.5cm}
\caption{Overview of {Ld-3DHS}. (a) The late-decoupled 3DHS branch uses a shared point cloud encoder with multiple late-decoupled decoders to perform hierarchical segmentation tasks, and leverages a coarse-to-fine guidance mechanism to balance task-specific learning and hierarchical consistency. Meanwhile, (b) the auxiliary discrimination branch introduces a novel semantic-prototype-based bi-branch supervision scheme, which employs contrastive learning to learn discriminative representations for individual classes, and then provides bi-directional supervision with the above 3DHS branch, thereby improving the segmentation ability on class-imbalance point clouds.}
\label{fig:pipeline}
\end{figure*}


To address these challenges, we propose \emph{Late-decoupled 3D Hierarchical Semantic Segmentation with Semantic Prototype Discrimination based Bi-branch Supervision}.
As shown in Figure~\ref{fig:pipeline}, our framework is named as Ld-3DHS driven by two insights.
First, we consider the multi-hierarchy segmentation task to be analogous to a multi-label classification task~\cite{sener2018multiobjective,chen2018gradnorm,yu2020pcgrad}, and thus using the shared parameters to fit different hierarchies' labels will lead to the overfitting at some levels and underfitting at others. This, in our context, creates the multi-hierarchy conflicts among the optimization objectives of different hierarchies.
Therefore, to mitigate this issue, we propose to establish a late-decoupled model for each hierarchical segmentation task, while multiple hierarchies share one point cloud feature extraction encoder to ensure a consistent information foundation.
Second, the class imbalance problem is expected to be constrained in each individual hierarchy due to the late-decoupled architecture.
Furthermore, motivated by contrastive learning \cite{chen2020simclr}, we introduce the auxiliary discriminative objectives for individual hierarchies, which separate class prototypes and guide the models to learn discriminative features for alleviating the class-imbalance segmentation tasks. In summary, our contributions are threefold:
\begin{itemize}
    \item We propose a novel late-decoupled 3DHS framework which employs multiple decoders with the coarse-to-fine hierarchical guidance and consistency, to alleviate the conflicts between underfitting and overfitting issues arising from multi-label learning in hierarchical segmentation tasks with shared model parameters.
    \item We introduce a 3DHS-oriented semantic prototype based bi-branch supervision method, that additionally learns class-wise discriminative point cloud features and performs mutual supervision between the auxiliary and 3DHS branches, to enhance the class-imbalance segmentation.
    \item Our approach achieves state-of-the-art 3DHS performance across multiple datasets and backbones, and can be integrated into previous methods to further improve their 3DHS performance.
\end{itemize}

\section{Related Work}\label{sec:relatedworks}

\subsection{Deep 3D Semantic Segmentation}
Recently, 3D semantic segmentation (3DS) has achieved significant progress with the development of deep models.
For example, PointNet~\cite{qi2017pointnet} is a pioneering work that directly learns point features with multi-layer perceptron instead of using handcrafted features and is widely employed to achieve deep 3DS.
Furthermore, PointCNN~\cite{xu2018spidercnn} leverages 3D convolutional networks to improve point cloud feature learning.
Motivated by the success of deep 3DS, many efforts have been made in this community and proposed a series of gradually developed methods, such as KPConv~\cite{thomas2019kpconv,thomas2024kpconvx}, MinkowskiNet~\cite{choy2019minkowski}, PVCNN~\cite{liu2019pvcnn}, DGCNN~\cite{wang2019dgcnn}, and Point Transformer~\cite{guo2021pct,zhao2021pointtransformer}.
Some work~\cite{hu2020randla,zhou2020cylinder3d,milioto2019rangenet} focuses on developing efficient methods to process large-scale point cloud scenes.
These advances are supported by large-scale datasets such as S3DIS~\cite{armeni2016s3dis}, Campus3D~\cite{li2020campus3d} and SensatUrban~\cite{hu2022sensaturban}.
However, existing 3DS methods usually ignore the hierarchical relationships in real-world scenes~\cite{li2020campus3d}.
They can only predict flat semantic labels and might suffer from performance degradation under class imbalance~\cite{peri2023towards,li2024classimbalanced}.
To address these issues, we propose a hierarchy-aware deep 3DS framework and consider mitigating the negative effect of class imbalance.

\subsection{3D Hierarchical Semantic Segmentation}
The concept of hierarchical segmentation, first explored in 2D image analysis to provide multi-hierarchy understanding, has recently been adapted to 3D point clouds.
Specifically, hierarchical semantic segmentation approaches~\cite{farabet2013learning,lin2017feature,zhao2017pspnet} show that combining coarse vision features with fine ones can substantially improve segmentation accuracy.
This observation highlights the importance of modeling semantic structures across multiple hierarchies for establishing 3D hierarchical semantic segmentation (3DHS).
To this end, Campus3D~\cite{li2020campus3d}, a large-scale outdoor point cloud benchmark with multi-hierarchy annotations, was introduced to support training 3DHS models.
Then, \citet{li2025deep_hierarchical_learning} proposed deep hierarchical learning to promote cross-hierarchy semantic consistency in 3DHS.
Although these 3DHS methods achieved progress, they still face the challenge of class imbalance.
Moreover, previous work has ignored the multi-hierarchy conflicts when optimizing deep 3DHS models.
In this paper, we propose a novel late-decoupled 3DHS framework to address these issues.

\subsection{Self-Supervised Learning}
To reduce reliance on costly annotations, self-supervised learning (SSL) has been widely explored for 3D vision.
For instance, PointContrast~\cite{xie2020pointcontrast} learns robust features by optimizing contrastive loss to align point features of different views.
Point-MAE~\cite{pang2022pointmae,pang2021masking,yu2022pointbert} reconstruct masked point cloud patches to train 3D representation learning models.
SSL-based 3D segmentation methods (e.g., GrowSP~\cite{zhang2023growsp}) show promising directions for label-efficient segmentation via superpoint growth and clustering, but they lack explicit hierarchical supervision and struggle to guarantee cross-hierarchy semantic coherence.
Recent work has started to address class imbalance by using prototype based methods~\cite{han2024subspace, han2024invspacenet} and class-imbalanced semi-supervised frameworks~\cite{li2024classimbalanced}.
These methods demonstrate that prototype guidance is effective for improving rare class performance.
Motivated by their success, in this work, we leverage contrastive learning and construct a self-supervision mechanism to improve model discrimination ability.

\section{Method}

\subsection{Problem Definition}
The 3D hierarchical semantic segmentation (3DHS) task is to predict hierarchical semantic labels for each point in a 3D point cloud scene.
Formally, given a point cloud $\mathbf{X} \in \mathbb{R}^{N \times 3}$, 3DHS methods learn a function $\mathcal{F}$ that predicts a set of labels $\{\hat y_i^{(1)},\hat y_i^{(2)}, \ldots, \hat y_i^{(H)}\}$ with hierarchical structures for each point $\mathbf{x}_i \in \mathbf{X}$, where $N$ denotes the point number and $H$ is the hierarchy number.
Concretely, the 3DHS model can be formulated as follows:
\begin{equation}\label{3DHS}
    \{\mathbf{Y}^{(h)}\}_{h=1}^H = \mathcal{F}_{\theta}(\mathbf{X}),
\end{equation}
where $\mathbf{Y}^{(h)} = [\mathbf{y}_1^{(h)};\mathbf{y}_2^{(h)};\dots;\mathbf{y}_N^{(h)}] \in \mathbb{R}^{N \times K^{(h)}}$, $\mathbf{y}_i^{(h)} \in \mathbb{R}^{K^{(h)}}$represents the soft classification results of the $i$-th point in the $h$-th hierarchy, $K^{(h)}$ is the class number.
$\theta$ are the model parameters that need to be trained.
As a result, $\hat{y}_i^{(h)} = \arg \max_{j} y_{i,j}^{(h)},~y_{i,j}^{(h)} \in \mathbf{y}_i^{(h)}$.

To address the multi-hierarchy conflict and class imbalance issue, we propose a novel dual-branch framework Ld-3DHS as shown in Figure~\ref{fig:pipeline}, which establishes a \emph{late-decoupled 3DHS branch} to avoid multi-hierarchy conflicts as well as trains an \emph{auxiliary discrimination branch} for mitigating class imbalance issue.
The following sections introduce the method details.

\subsection{Late-Decoupled 3DHS with Coarse-to-Fine Hierarchical Guidance and Consistency}

\noindent\textbf{Late-decoupled 3DHS branch.}
As shown in Figure~\ref{fig:pipeline}(a), we propose the late-decoupled 3DHS branch to avoid multi-hierarchy conflicts.
Specifically, our method improves the conventional formulation of Eq.~(\ref{3DHS}) as follows:
\begin{equation}\label{eq:prediction}
    \{\mathbf{Y}^{(h)}\}_{h=1}^H = \{\mathcal{G}^{(h)}_{\delta^{(h)}}(\mathbf{Z})\}_{h=1}^H = \{\mathcal{G}^{(h)}_{\delta^{(h)}}(\mathcal{E}_{\theta}(\mathbf{X}))\}_{h=1}^H,
\end{equation}
where $\mathbf{Z} = \mathcal{E}_{\theta}(\mathbf{X})$ denotes the early point cloud feature extraction, while $\mathbf{Y}^{(h)} = \mathcal{G}^{(h)}_{\delta^{(h)}}(\mathbf{Z})$ indicates that we establish the late-decoupled 3DHS model with individual parameter $\delta^{(h)}$ to achieve hierarchy-wise semantic segmentation.

This design makes the foundational knowledge in the shared encoder $\mathcal{E}_{\theta}$ transfers efficiently to all hierarchies, while ensuring that the decoupled decoders $\{\mathcal{G}^{(h)}_{\delta^{(h)}}\}^H_{h=1}$ can mitigate the negative conflicts among diverse multi-hierarchy semantic segmentation tasks.

\vspace{+0.05cm}
\noindent\textbf{Coarse-to-fine hierarchical guidance.}
To fully utilize the multi-hierarchy discriminability, we propose the coarse-to-fine hierarchical guidance which fuses the semantic prediction of one hierarchy with the features of the next hierarchy, enabling their information flow and mutual enhancement.

To be specific, we use $\mathbf{H}^{(h-1)}$ to represent the middle feature between $\mathbf{Y}^{(h-1)}$ and $\mathbf{Z}$.
Then, the coarse-to-fine hierarchical guidance module can be formulated as follows:
\begin{equation}\label{eq:feature_fusion}
    {\hat{\mathbf{H}}^{(h)}} = \text{MLP}\left( \left[ \mathbf{H}^{(h)} \oplus \alpha \cdot \text{MLP}(\mathbf{Y}^{(h-1)}) \right] \right),
\end{equation}
where \text{MLP} denotes the multi-layer perceptron, $\oplus$ denotes the concatenate operation, and $\alpha$ is a trade-off parameter to balance the influence between two hierarchies.
Then, the semantic prediction for the $h$-th hierarchy is obtained by
\begin{equation}\label{eq:semantic_prior}
    \mathbf{Y}^{(h)} = \text{Classifier}(\hat{\mathbf{H}}^{(h)}).
\end{equation}
the \text{classifier} module projects the fused feature $\hat{\mathbf{H}}^{(h)}$ to the probabilistic prediction $\mathbf{Y}^{(h)}$.
In this way, the late-decoupled 3DHS models are trained by cross-entropy loss:
\begin{equation}\label{eq:prediction_loss}
    \mathcal{L}_{\text{ces}} = \sum_{h=1}^{H} \mathcal{L}^{(h)} = - \sum_{h=1}^{H} \frac{1}{N} \sum_{i=1}^{N} \sum_{j=1}^{K^{(h)}} \hat{y}_{i,j}^{(h)} \log({y}_{i,j}^{(h)}).
\end{equation}
$\hat{y}_{i,j}^{(h)}$ is the one-hot ground-truth label ($1$ if point $i$ belongs to class $j$, $0$ otherwise), and $y_{i,j}^{(h)} \in \mathbf{Y}^{(h)}$ is the model's output probability for point $i$ on class $j$.

\vspace{+0.05cm}
\noindent\textbf{Cross-hierarchical consistency.}
To take advantage of the parent-child structures~\cite{Athanasopoulos2023Forecast,li2025deep_hierarchical_learning} within hierarchical labels in Eq.~(\ref{eq:prediction_loss}), we further establish the cross-hierarchical consistency loss for optimizing the late-decoupled 3DHS models.
Concretely, we define a mapping matrix $\mathbf{A}^{(h,h-1)} \in \{0,1\}^{K^{(h)} \times K^{(h-1)}}$ which indicates the parent-child relationship between the $(h-1)$ and $h$-th hierarchies.
The mapping matrix can convert coarse-grained labels into fine-grained labels, and its formal definition can be written as
\begin{equation}
    \min_{\mathbf{A}}|{\hat{\mathbf{Y}}}^{(h)} -  {\hat{\mathbf{Y}}}^{(h-1)} (\mathbf{A}^{(h,h-1)})^T|,
    \label{eq:a}
\end{equation}
where $(\cdot)^T$ is the matrix transposition operation, ${\hat{\mathbf{Y}}}^{(h)} \in \{0,1\}^{N\times K^{(h)}}$ and ${\hat{\mathbf{Y}}}^{(h-1) } \in \{0,1\}^{N\times K^{(h-1)}}$ denote the one-hot label matrices.
Then, the following loss is employed to achieve our cross-hierarchical consistency objective:
\begin{equation}\label{eq:hc}
    \mathcal{L}_{\text{chc}} = \frac{1}{N} \sum_{i=1}^{N} \sum_{h=2}^{H} \left \| {\mathbf{y}}^{(h)}_{i} - \mathbf{A}^{(h,h-1)} {\mathbf{y}}^{(h-1)}_{i} \right \|_2^2.
\end{equation}
where ${\mathbf{y}}^{(h)}_{i}$ is the prediction vector for the point $i$ at the hierarchy level $h$ with a shape of $K^{(h)} \times 1$.

Finally, the objective for training our late-decoupled 3DHS branch combines the cross-entropy and the cross-hierarchical consistency losses as follows:
\begin{equation}\label{eq:main_loss}
    \mathcal{L}_{\text{late-3DHS}} = \mathcal{L}_{\text{ces}} + \mathcal{L}_{\text{chc}}.
\end{equation}

\subsection{Auxiliary Discrimination with Semantic Prototype based Bi-branch Supervision}

\noindent\textbf{Auxiliary discrimination branch.}
In this paper, we introduce the auxiliary discrimination branch to mitigate the class-imbalance issue in 3DHS as shown in Figure~\ref{fig:pipeline}(b).
Specifically, in the training process, we divide the complete scene point cloud $\mathbf{X}$ into multiple class-wise point cloud $\{\mathbf{X}_i^{(h)}\}_{i=1}^{K^{(h)}}$, where $\mathbf{X}_i^{(h)}$ denotes the $i$-th class data in the $h$-th hierarchy. Then, we employ contrastive learning to learn class-wise discriminative features, which are further used to construct semantic prototypes to provide helpful supervision for improving our late-decoupled 3DHS branch.

\vspace{+0.05cm}
\noindent\textbf{Contrastive discriminative features.}
Given $\{\mathbf{X}^{(h,c)}\}_{c=1}^{K^{(h)}}$, we first utilize the point cloud encoder to extract class-wise features and stack a MLP projection head to obtain contrastive features $\{\mathbf{F}^{(h,c)}\}_{c=1}^{K^{(h)}}$.
Then, we minimize the supervised contrastive loss~\cite{khosla2020supervised} to achieve the feature discrimination objective:
\begin{equation}\label{eq:con_loss}
\mathcal{L}_{\text{con}}^{(h)} = -\mathbb{E}_{s^+ \in \mathcal{P}^{(h)}} \left[ s^+ - \log \sum\nolimits_{s^- \in \mathcal{N}^{(h)}} e^{s^-} \right],
\end{equation}
where $\mathcal{P}^{(h)}$ and $\mathcal{N}^{(h)}$ denote the sets of positive and negative pairs, respectively.
If the features $\mathbf{f}_i^{(h,c_1)} \in \mathbf{F}^{(h,c_1)}$ and $\mathbf{f}_j^{(h,c_2)} \in \mathbf{F}^{(h,c_2)}$ come from the same class (i.e., $c_1 = c_2$) in the $h$-th hierarchy, their feature pair belongs to $\mathcal{P}^{(h)}$, otherwise the pair belongs to $\mathcal{N}^{(h)}$.
$s^+$ and $s^-$ measure the feature distance computed by cosine similarity.

This discrimination objective is expected to pull the intra-class features closer together while pushing inter-class features further apart, thereby learning discriminative point cloud features for individual classes.

\vspace{+0.05cm}
\noindent\textbf{Semantic-prototype-based bi-branch supervision.}
Given the class-wise discriminative features $\{\mathbf{F}^{(h,c)}\}^{K^{(h)}}_{c=1}$ in the auxiliary discrimination branch,
we use the same manner to divide the middle features $\mathbf{H}^{(h)}$ into class-wise middle features $\{\mathbf{H}^{(h,c)}\}_{c=1}^{K^{(h)}}$ in the late-decoupled 3DHS branch.
Then, we propose the semantic-prototype-based bi-branch supervision mechanism to promote 3DHS tasks.
Formally, for the $h$-th hierarchy, we compute semantic prototypes on both branches using the feature expectation of each class:
\begin{equation}\label{eq:prototype_generation_h}
    \mathbf{p}_{\text{3D}}^{(h,c)} = \mathbb{E}_i[\mathbf{h}_i^{(h,c)}], ~{\mathbf{p}}_{\text{aux}}^{(h,c)} = \mathbb{E}_i[\mathbf{f}_i^{(h,c)}],
\end{equation}
where $\mathbf{h}_i^{(h,c)} \in \mathbf{H}^{(h,c)}$, $\mathbf{f}_i^{(h,c)} \in \mathbf{F}^{(h,c)}$ denote the $i$-th sample features in the $c$-th class.
$\mathbf{p}_{\text{3D}}^{(h,c)}$ and $\mathbf{p}_{\text{aux}}^{(h,c)}$ are defined as the semantic prototypes that extract the general class information from the 3DHS and auxiliary branches. Furthermore, we leverage the two prototypes to mutually supervise two branches by:
{\small\begin{equation}\label{eq:bs_loss_final}
\begin{aligned}\mathcal{L}_{\text{bis}}^{(h)} 
&= \sum_{c} \frac{1}{N^{(h,c)}} \sum_{i} 
\Big[
    \mathrm{Smooth}_{L1}\big(\mathbf{p}_{\text{3D}}^{(h,c)} - \mathbf{f}^{(h,c)}_{i}\big) \\
&\qquad\qquad\quad
    ~~~~~~~~+\mathrm{Smooth}_{L1}\big(\mathbf{p}_{\text{aux}}^{(h,c)} - \mathbf{h}^{(h,c)}_{i}\big)
\Big],
\end{aligned}
\end{equation}}
where $N^{(h,c)}$ is the feature number of the $c$-th class in the $h$-th hierarchy.
$\mathrm{Smooth}_{L1}$ denotes the smooth L1 loss~\cite{girshick2015fastrcnn} commonly employed to increase the robustness to outliers, which satisfies that
$\mathrm{Smooth}_{L1}(x) = 0.5 x^2,~\text{if } |x| < 1,\text{~otherwise~~}\mathrm{Smooth}_{L1}(x) =|x| - 0.5$.

In our auxiliary discrimination branch, the contrastive loss $\mathcal{L}_{\text{con}}^{(h)}$ and the bi-branch supervision loss $\mathcal{L}_{\text{bis}}^{(h)}$ are combined for joint optimization:
\begin{equation}
\mathcal{L}_{\text{aux}}^{(h)} = \mathcal{L}_{\text{con}}^{(h)} + \mathcal{L}_{\text{bis}}^{(h)}.
\label{eq:aux_loss_final}
\end{equation}
Overall, the entire loss in our method is formulated as:
\begin{equation}
    \mathcal{L}_{\text{total}} = \mathcal{L}_{\text{late-3DHS}} + \lambda \sum\nolimits_{h=1}^{H} \mathcal{L}_{\text{aux}}^{(h)},
    \label{eq:total_loss}
\end{equation}
where $\lambda$ is a hyper-parameter to balance the two branches' training.
In this way, our model is expected to leverage the class-wise information in the auxiliary discrimination branch to promote the hierarchical semantic segmentation tasks in the late-decoupled 3DHS branch.

\begin{table*}[!t]
\centering
\caption{\textbf{3D Hierarchical Semantic Segmentation on Three Datasets.} Results of mIoU (\%) are reported. Our method (Ld-3DHS) is compared against MTHS and DHL across various backbones. The improvement of Ld-3DHS over the best prior method is shown in \textcolor{mydarkred}{red}.}
\label{tab:combined_results_final_revised}
\vspace{-0.2cm}
\small
\resizebox{\linewidth}{!}{
\begin{tabular}
{l|l|cccc|ccc|ccc}
\toprule[2pt]
\multirow{2}{*}{\textbf{Backbone}} & \multirow{2}{*}{\textbf{Method}} & \multicolumn{4}{c|}{\textbf{Campus3D}} & \multicolumn{3}{c|}{\textbf{S3DIS-H}} & \multicolumn{3}{c}{\textbf{SensatUrban-H}} \\
\cline{3-12}
& & \textbf{Avg. mIoU} & \textbf{L0} & \textbf{L1} & \textbf{L2} & \textbf{Avg. mIoU} & \textbf{L0} & \textbf{L1} & \textbf{Avg. mIoU} & \textbf{L0} & \textbf{L1} \\
\hline
\multirow{3}{*}{PointNet++}
& MTHS~\cite{li2020campus3d} & 58.80 & 92.90 & 42.50 & 41.00 & 62.74 & 66.95 & 58.52 & 47.20 & 51.89 & 42.51 \\
& DHL~\cite{li2025deep_hierarchical_learning}  & 62.56 & 91.68 & \textbf{54.89} & 41.12 & 63.05 & 67.45 & 58.65 & 48.20 & \textbf{57.20} & 39.20 \\
& \textbf{Ld-3DHS (ours)} & \cellcolor{gray!15}\textbf{63.28} (\textcolor{mydarkred}{+0.72}) & \cellcolor{gray!15}\textbf{93.02} & \cellcolor{gray!15}52.22 & \cellcolor{gray!15}\textbf{44.61} & \cellcolor{gray!15}\textbf{66.43} (\textcolor{mydarkred}{+3.38}) & \cellcolor{gray!15}\textbf{70.06} & \cellcolor{gray!15}\textbf{62.80} & \cellcolor{gray!15}\textbf{49.73} (\textcolor{mydarkred}{+1.53}) & \cellcolor{gray!15}54.08 & \cellcolor{gray!15}\textbf{44.42} \\
\hline
\multirow{3}{*}{Point TF v2}
& MTHS~\cite{li2020campus3d} & 63.85 & 92.40 & 56.14 & 43.01 & 74.90 & 78.56 & 71.23 & 49.27 & 54.22 & 44.32 \\
& DHL~\cite{li2025deep_hierarchical_learning}  & 65.80 & 92.72 & 60.36 & 44.31 & 75.76 & 79.66 & 71.86 & 50.62 & 55.62 & 45.62 \\
& \textbf{Ld-3DHS (ours)} & \cellcolor{gray!15}\textbf{66.87} (\textcolor{mydarkred}{+1.07}) & \cellcolor{gray!15}\textbf{93.05} & \cellcolor{gray!15}\textbf{61.85} & \cellcolor{gray!15}\textbf{45.71} & \cellcolor{gray!15}\textbf{76.71} (\textcolor{mydarkred}{+0.95}) & \cellcolor{gray!15}\textbf{80.30} & \cellcolor{gray!15}\textbf{73.12} & \cellcolor{gray!15}\textbf{53.56} (\textcolor{mydarkred}{+2.94}) & \cellcolor{gray!15}\textbf{58.98} & \cellcolor{gray!15}\textbf{48.14} \\
\hline
\multirow{3}{*}{Point TF v3}
& MTHS~\cite{li2020campus3d} & 64.96 & 92.90 & 57.83 & 44.15 & 71.54 & 75.59 & 67.48 & 39.07 & 42.90 & 35.24 \\
& DHL~\cite{li2025deep_hierarchical_learning}  & 62.90 & 91.20 & 55.50 & 42.00 & 72.83 & 76.32 & 69.34 & 39.41 & 43.29 & 35.53 \\
& \textbf{Ld-3DHS (ours)} & \cellcolor{gray!15}\textbf{66.40} (\textcolor{mydarkred}{+1.44}) & \cellcolor{gray!15}\textbf{93.40} & \cellcolor{gray!15}\textbf{60.50} & \cellcolor{gray!15}\textbf{45.30} & \cellcolor{gray!15}\textbf{74.50} (\textcolor{mydarkred}{+1.67}) & \cellcolor{gray!15}\textbf{78.30} & \cellcolor{gray!15}\textbf{70.70} & \cellcolor{gray!15}\textbf{40.43} (\textcolor{mydarkred}{+1.02}) & \cellcolor{gray!15}\textbf{44.42} & \cellcolor{gray!15}\textbf{36.43} \\
\bottomrule[2pt]
\end{tabular}
}
\end{table*}

\section{Experiments}
This section shows extensive experiments to evaluate the effectiveness of our proposed framework.
We first detail the experimental setup in Sec.~\ref{exp_setup}. 
After that, quantitative comparisons against state-of-the-art methods are demonstrated in Sec.~\ref{quantitative}, followed by qualitative visualizations in Sec.~\ref{qualitative}. 
Finally, we conduct in-depth ablation studies in Sec.~\ref{sec:ablation} to validate the key components of our method.

\subsection{Experimental Setup}\label{exp_setup}

\noindent\textbf{Datasets.} 
We conduct experiments on three datasets featuring hierarchical annotations.
\textbf{1) Campus3D:} The Campus3D~\cite{li2020campus3d} dataset natively provides 5 semantic hierarchies. In our experiments, we select the representative 1-st, 3-rd, and 5-th hierarchies to form a 3-hierarchy evaluation benchmark. Following~\cite{li2025deep_hierarchical_learning}, to handle extreme sparsity at the finest granularity, the three least common classes are merged into a single `miscellaneous' category. We follow the standard evaluation process, i.e., training on area FASS, YIH, RA, UCC, validating on PGP, and testing on FOE.
\textbf{2) S3DIS-H:} The Stanford 3D Indoor Spaces (S3DIS~\cite{armeni2016s3dis}) dataset consists of 6 large indoor areas with 13 semantic categories. As this dataset lacks native hierarchical labels, we constructed a 2-hierarchy, which we term S3DIS-H, by following the automated pipeline~\cite{li2025deep_hierarchical_learning}. We adhere to the standard evaluation protocol, training on Areas 1-4, 6 and testing on Area 5.
\textbf{3) SensatUrban-H:} The SensatUrban-H dataset is a 2-hierarchy large-scale urban benchmark. This version was constructed by \cite{li2025deep_hierarchical_learning} from the original SensatUrban~\cite{hu2021sensaturban} dataset, and we directly utilize their publicly released version, which partitions the dataset's blocks from the two cities, i.e., Birmingham and Cambridge, into distinct training, validation, and test sets.

\vspace{+0.05cm}
\noindent\textbf{Comparison baselines.} We compare our method against two state-of-the-art 3DHS methods: MTHS~\cite{li2020campus3d} and DHL~\cite{li2025deep_hierarchical_learning}.
To ensure a fair comparison that isolates the contribution of our methodology from architectural differences, we evaluate all methods on the same backbones including PointNet++~\cite{qi2017pointnetplusplus}, Point Transformer v2~\cite{wu2022point} (Point TF v2), and Point Transformer v3~\cite{wu2024ptv3} (Point TF v3).
This allows a comprehensive evaluation of our method's effectiveness.

\vspace{+0.05cm}
\noindent\textbf{Evaluation metrics.} The primary evaluation metric for all experiments is the {mean Intersection over Union (mIoU)}. Crucially, to provide a comprehensive assessment of 3DHS methods, we report the mIoU score for each hierarchy separately and also report the average mIoU (Avg.mIOU) across all hierarchies for an overall performance summary.

\vspace{+0.05cm}
\noindent\textbf{Implementation details.}
We employ an adaptive decoder strategy to balance performance and efficiency. To be specific, for the lightweight backbone PointNet++, we use multiple copies of the original decoder to implement the late-decoupled decoders.
For Point Transformer v2 and v3, multiple lightweight MLPs are used as late-decoupled decoders to achieve efficient 3DHS.
All experiments are conducted on a single NVIDIA RTX 4090 GPU (24GB) using PyTorch v2.0.0 with CUDA 11.8.
Models were trained for 100 epochs using the AdamW optimizer with betas of (0.9, 0.999) and a weight decay of 1e-4.
The learning rate is managed by a cosine annealing scheduler, starting at 0.01 and decaying to 1.0e-5.
The batch size is set to 12 for S3DIS-H and 64 for Campus3D and SensatUrban-H.
To enhance model robustness and prevent over-fitting, we implement a series of online data augmentations including random scaling within the $[0.9, 1.1]$ range, random point jittering with a Gaussian noise of $\sigma=0.005$, and a 20\% probability of color dropping.
In all our experiments, all hyper-parameters for weighting the different loss components are set to $1.0$.

\begin{table}[!t]
\centering
\caption{Performance gain (Avg. mIoU) before and after applying our Ld-3DHS to MTHS and DHL using PointNet++.}
\label{tab:pg1}
\vspace{-0.2cm}
\renewcommand{\arraystretch}{1.2} 
\resizebox{\columnwidth}{!}{%
\begin{tabular}{l|ccc}
\toprule[2pt]
\multirow{2}{*}{\textbf{Method}} & \multicolumn{3}{c}{\textbf{Dataset}} \\
\cline{2-4}
& \textbf{Campus3D} & \textbf{S3DIS-H} & \textbf{SensatUrban-H} \\
\hline
MTHS (baseline) & 58.80 & 62.74 & 47.20 \\
\cellcolor{gray!15}\textbf{MTHS + Ld-3DHS} & \cellcolor{gray!15}\textbf{59.96} \textcolor{mydarkred}{\footnotesize(+1.16)} & \cellcolor{gray!15}\textbf{63.93} \textcolor{mydarkred}{\footnotesize(+1.19)} & \cellcolor{gray!15}\textbf{49.05} \textcolor{mydarkred}{\footnotesize(+1.85)} \\
DHL (baseline) & 62.56 & 63.05 & 48.20 \\
\cellcolor{gray!15}\textbf{DHL + Ld-3DHS} & \cellcolor{gray!15}\textbf{64.29} \textcolor{mydarkred}{\footnotesize(+1.73)} & \cellcolor{gray!15}\textbf{63.59} \textcolor{mydarkred}{\footnotesize(+0.54)} & \cellcolor{gray!15}\textbf{49.12} \textcolor{mydarkred}{\footnotesize(+0.92)} \\
\bottomrule[2pt]
\end{tabular}
}
\end{table}

\begin{table}[!t]
\centering
\caption{Performance gain (Avg. mIoU) before and after applying our Ld-3DHS to MTHS and DHL using Point Transformer v2.}
\label{tab:pg2}
\vspace{-0.2cm}
\renewcommand{\arraystretch}{1.2} 
\resizebox{\columnwidth}{!}{%
\begin{tabular}{l|ccc}
\toprule[2pt]
\multirow{2}{*}{\textbf{Method}} & \multicolumn{3}{c}{\textbf{Dataset}} \\
\cline{2-4}
& \textbf{Campus3D} & \textbf{S3DIS-H} & \textbf{SensatUrban-H} \\
\hline
MTHS (baseline) & 63.85 & 74.90 & 49.27 \\
\cellcolor{gray!15}\textbf{MTHS + Ld-3DHS} & \cellcolor{gray!15}\textbf{66.38} \textcolor{mydarkred}{\footnotesize(+2.53)} & \cellcolor{gray!15}\textbf{76.40} \textcolor{mydarkred}{\footnotesize(+1.50)} & \cellcolor{gray!15}\textbf{51.07} \textcolor{mydarkred}{\footnotesize(+1.80)} \\
DHL (baseline) & 65.80 & 75.76 & 50.62 \\
\cellcolor{gray!15}\textbf{DHL + Ld-3DHS} & \cellcolor{gray!15}\textbf{66.94} \textcolor{mydarkred}{\footnotesize(+1.14)} & \cellcolor{gray!15}\textbf{76.76} \textcolor{mydarkred}{\footnotesize(+1.00)} & \cellcolor{gray!15}\textbf{53.61} \textcolor{mydarkred}{\footnotesize(+2.99)} \\
\bottomrule[2pt]
\end{tabular}
}
\end{table}
\begin{table}[!t]
\centering
\caption{Performance gain (Avg. mIoU) before and after applying our Ld-3DHS to MTHS and DHL using Point Transformer v3.}
\label{tab:pg3}
\vspace{-0.2cm}
\renewcommand{\arraystretch}{1.2} 
\resizebox{\columnwidth}{!}{%
\begin{tabular}{l|ccc}
\toprule[2pt]
\multirow{2}{*}{\textbf{Method}} & \multicolumn{3}{c}{\textbf{Dataset}} \\
\cline{2-4}
& \textbf{Campus3D} & \textbf{S3DIS-H} & \textbf{SensatUrban-H} \\
\hline
MTHS (baseline) & 64.96 & 71.54 & 39.07 \\
\cellcolor{gray!15}\textbf{MTHS + Ld-3DHS} & \cellcolor{gray!15}\textbf{66.12} \textcolor{mydarkred}{\footnotesize(+1.16)} & \cellcolor{gray!15}\textbf{74.20} \textcolor{mydarkred}{\footnotesize(+2.66)} & \cellcolor{gray!15}\textbf{40.12} \textcolor{mydarkred}{\footnotesize(+1.05)} \\
DHL (baseline) & 62.90 & 72.83 & 39.41 \\
\cellcolor{gray!15}\textbf{DHL + Ld-3DHS} & \cellcolor{gray!15}\textbf{64.67} \textcolor{mydarkred}{\footnotesize(+1.77)} & \cellcolor{gray!15}\textbf{75.34} \textcolor{mydarkred}{\footnotesize(+2.51)} & \cellcolor{gray!15}\textbf{40.44} \textcolor{mydarkred}{\footnotesize(+1.03)} \\
\bottomrule[2pt]
\end{tabular}
}
\end{table}

\begin{figure}[!t]
\centering
  \begin{subfigure}{\linewidth}
    \includegraphics[width=\linewidth]{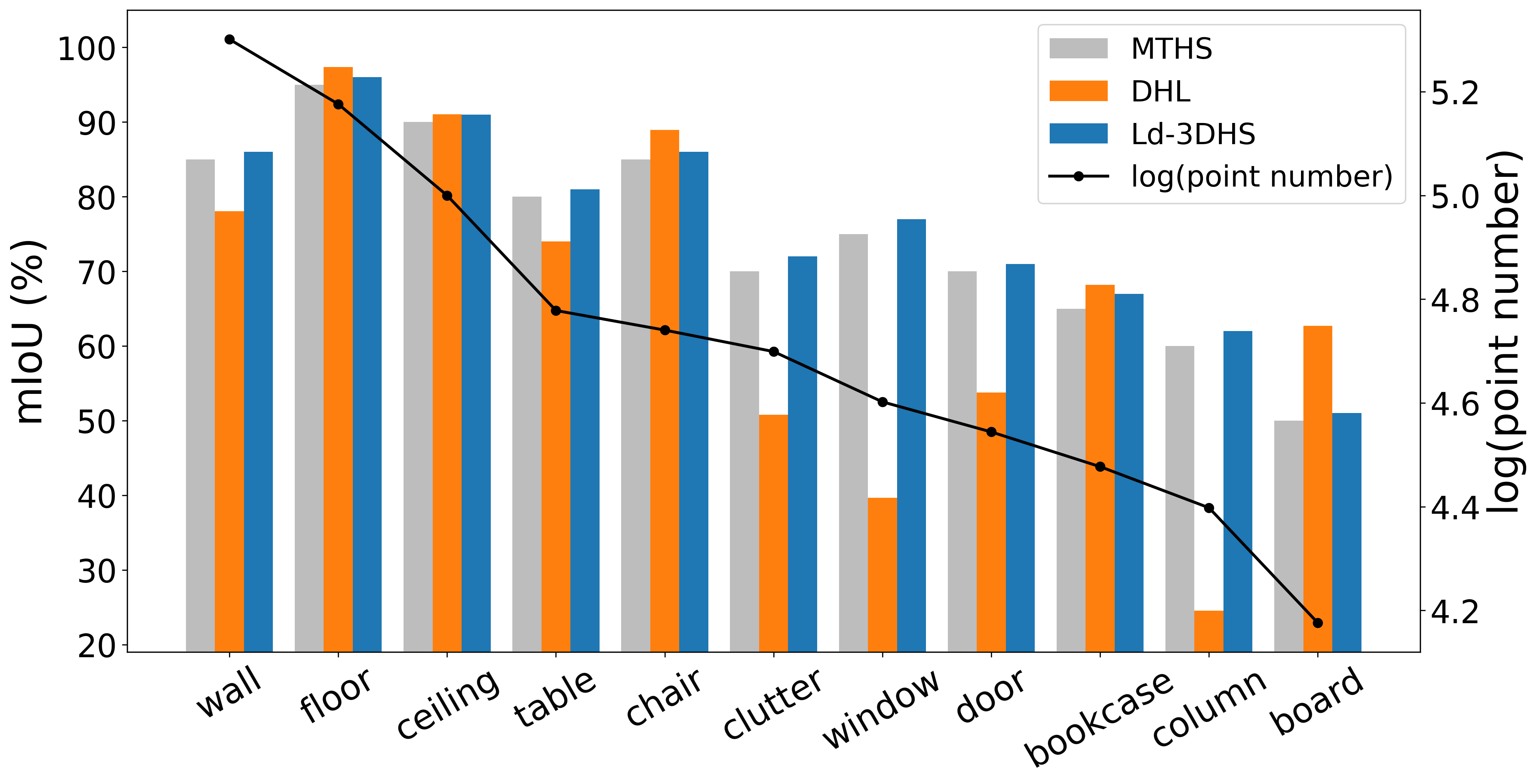}
  \end{subfigure}
\vspace{-0.7cm}
\caption{Per-class segmentation performance comparison of three methods on S3DIS-H dataset. The mIoU values of the L1 hierarchy on S3DIS-H dataset are reported, and the black line records the log-scale point number for individual classes.}
\label{fig:per_class_comparison_vertical}
\end{figure}
\begin{figure}[!t]
    \centering
    \includegraphics[width=0.99\linewidth]{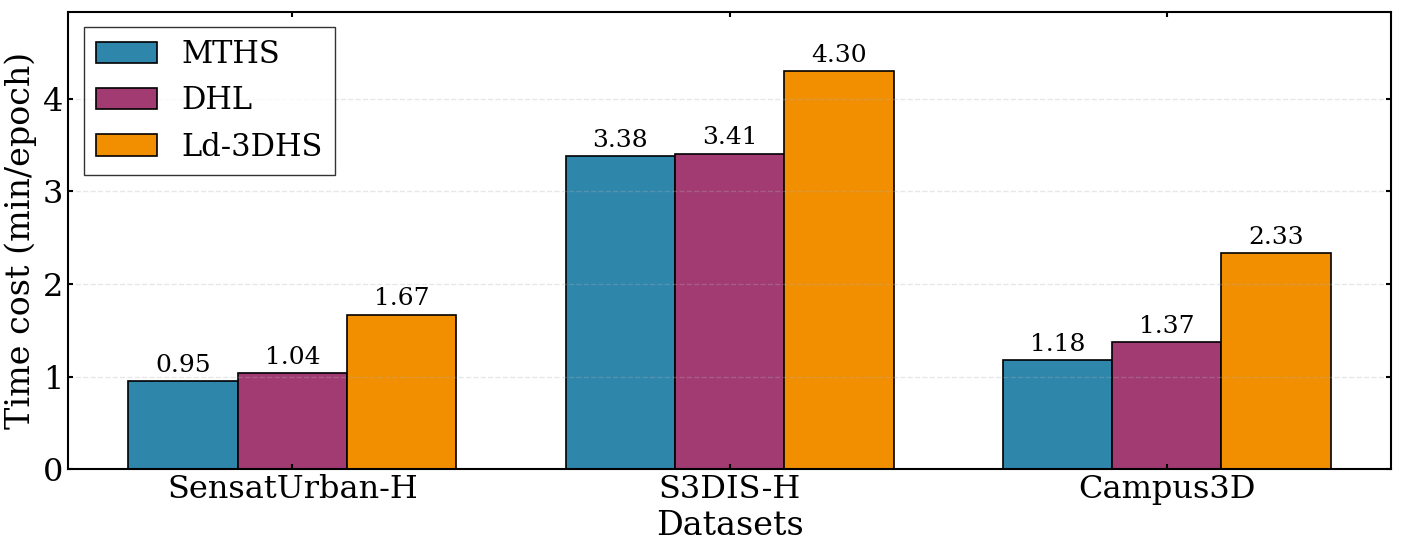}
    \vspace{-0.3cm}
    \caption{Time cost comparison of three 3DHS methods.}
    \label{fig:time}
\end{figure}

\subsection{Quantitative Comparison.}\label{quantitative}


\noindent\textbf{Comparison experiments.} We comprehensively compare our method Ld-3DHS with the SOTA methods MTHS and DHL across three datasets based on three backbones.
The results are listed in Table~\ref{tab:combined_results_final_revised}, from which we can observe that:
1) Our Ld-3DHS consistently achieves superior 3DHS performance across all tested datasets.
For instance, with the same backbone PointNet++, the Avg.mIoU of Ld-3DHS surpasses that of DHL by \textbf{3.38\%} on indoor benchmark S3DIS-H and \textbf{1.53\%} on outdoor benchmark SensatUrban-H.
2) Our Ld-3DHS is compatible with different backbones and has achieved new SOTA results.
For example, on Campus3D, Ld-3DHS outperforms the highly competitive DHL by \textbf{0.72\%}, \textbf{1.07\%}, and \textbf{1.44\%} Avg.mIoU using PointNet++, Point TF v2, and Point TF v3, respectively.
These improvements indicate that our Ld-3DHS achieves better segmentation performance at multiple hierarchies.

\vspace{+0.05cm}
\noindent\textbf{Add-on experiments.} 
In Tables~\ref{tab:pg1}, \ref{tab:pg2}, and~\ref{tab:pg3}, we conduct add-on experiments by applying our Ld-3DHS to previous methods MTHS and DHL.
Specifically, we employ the model structure and loss function of our Ld-3dHS to methods MTHS and DHL to reconstruct and then re-run their models.
The results indicate that Ld-3DHS can excel as an auxiliary module to boost the performance of previous methods.
For example, on SensatUrban-H dataset and PointNet++ backbone, Ld-3DHS yields \textbf{+1.85\%} improvement of Avg.mIoU when it is integrated with the MTHS baseline.
On Campus3D dataset and Point TF v3 backbone, our Ld-3DHS contributes to an Avg.mIoU gain of \textbf{+1.77\%}.
As a consequence, applying our Ld-3DHS consistently improves the hierarchical segmentation results across different datasets, backbones, and baseline methods, thus further validating our method's effectiveness.

\subsection{Qualitative Results}
\label{qualitative}

\begin{figure*}[!t]
\centering
\begin{overpic}
[width=\linewidth]
{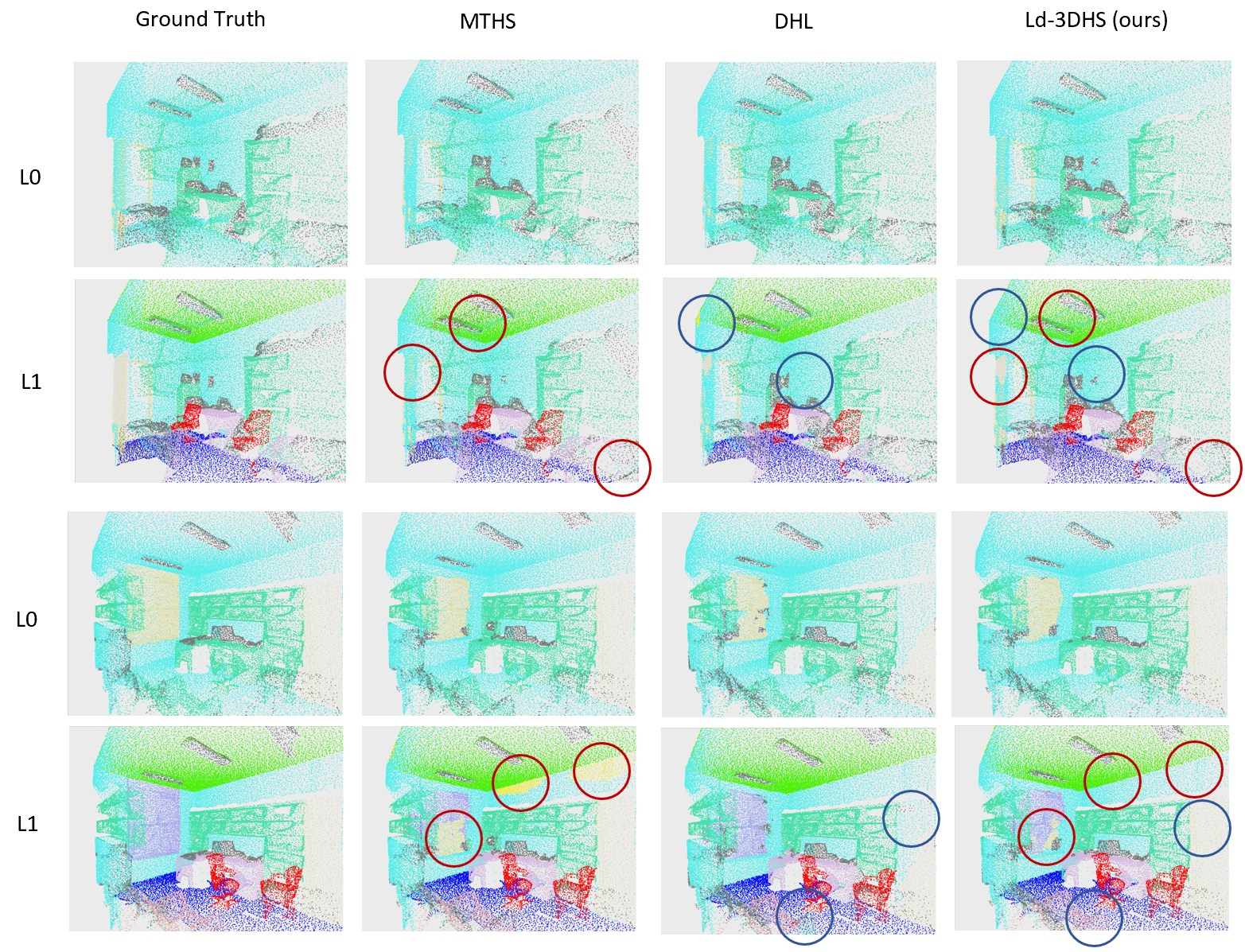}
\end{overpic}
\vspace{-0.7cm}
\caption{
Case comparison of three 3D hierarchical semantic segmentation methods (MTHS~\cite{li2020campus3d}, DHL~\cite{li2025deep_hierarchical_learning}, and our Ld-3DHS) on testing samples of S3DIS-H.
L0 and L1 represent the two semantic hierarchies in the indoor point cloud scenes.
}
\label{fig:gallery}
\vspace{-0.3cm}
\end{figure*}

\noindent\textbf{Segmentation performance $vs.$ point number.}
In Figure~\ref{fig:per_class_comparison_vertical}, we visualize the segmentation performance on each class and the corresponding point number on the finest hierarchy of S3DIS-H.
Compared with MTHS and DHL, our method achieves comparable performance on the majority classes, such as \emph{wall}, \emph{floor}, \emph{ceiling}, \emph{table}, and \emph{chair}.
Notably, our method achieves significant performance improvement on the minority classes, such as \emph{clutter}, \emph{window}, \emph{door}, and \emph{column}.
These trends in Figure~\ref{fig:per_class_comparison_vertical} are obvious and they demonstrate the effectiveness of our proposed method in addressing the class imbalance issue in 3DHS tasks.


\noindent\textbf{Time cost comparison for 3DHS.}
To evaluate the time cost of MTHS, DHL, and our Ld-3DHS, we record their average time per epoch in training stages on the same device.
The results on three datasets using PointNet++ backbone are shown in Figure~\ref{fig:time} which indicates that the time consumption of our method is slightly higher than that of the two comparison methods.
This is mainly because our method needs to train the auxiliary discrimination branch to improve segmentation performance.
Overall, our method achieves a more robust 3DHS at an affordable time cost.


\noindent\textbf{Case comparison for 3DHS.}
Figure~\ref{fig:gallery} provides a qualitative comparison of our method against MTHS and DHL on the challenging indoor benchmark S3DIS-H.
It shows that all three methods have satisfactory performance at the coarse-grained point cloud segmentation, e.g., at the L0 hierarchy as shown in the visualization cases.
However, we can observe the failure cases at the fine-grained point cloud segmentation (L1 hierarchy) of previous 3DHS methods.
For example, as indicated by the red circles in MTHS column, MTHS mis-segments the points of wall and window compared to the ground truth.
As indicated by the blue circles in DHL column, DHL mis-segments the points at the corner and boundary compared to the ground truth.
In contrast, our method successfully overcomes the challenging segmentation tasks for those point clouds.
As highlighted in the red and blue circles in the last column, our method correctly segments the points (which are wrongly segmented by MTHS and DHL) compared with the ground truth, indicating its ability to perform fine-grained point cloud segmentation. 
These visualization results verified the effectiveness of our method again.






\subsection{Ablation Studies}
\label{sec:ablation}

\begin{table}[!t]
\centering
\caption{\textbf{Ablation study on model structure.} Results of Avg. mIoU (\%) are reported. LDF: late-decoupled framework, CFG: coarse-to-fine guidance, ADB: auxiliary discrimination branch.}\label{ta}
\vspace{-0.2cm}
\resizebox{\linewidth}{!}{ 
\begin{tabular}{cccccc}
\toprule[2pt]
\multicolumn{3}{c}{\textbf{Model variant}} & \multicolumn{3}{c}{\textbf{Dataset}} \\
\cmidrule(r){1-3} \cmidrule(r){4-6} 
\multirow{1}{*}{\textbf{LDF}} & \multirow{1}{*}{\textbf{CFG}} & \multirow{1}{*}{\textbf{ADB}} & \multirow{1}{*}{\textbf{Campus3D}} & \multirow{1}{*}{\textbf{S3DIS-H}} & \multirow{1}{*}{\textbf{SensatUrban-H}} \\
\midrule
\rowcolor{gray!10} \ding{51} & \ding{51} & \ding{51} & 63.28 & 66.43 & 49.25 \\
\ding{55} & \ding{51} & \ding{51} & 59.22 \textcolor{mydarkgreen}{\footnotesize(-4.06)} & 62.17 \textcolor{mydarkgreen}{\footnotesize(-4.26)} & 47.74 \textcolor{mydarkgreen}{\footnotesize(-1.51)} \\
\ding{51} & \ding{55} & \ding{51} & 59.96 \textcolor{mydarkgreen}{\footnotesize(-3.32)} & 63.93 \textcolor{mydarkgreen}{\footnotesize(-2.50)} & 49.05 \textcolor{mydarkgreen}{\footnotesize(-0.20)}\\
\ding{51} & \ding{51} & \ding{55} & 61.95 \textcolor{mydarkgreen}{\footnotesize(-1.33)} & 62.98 \textcolor{mydarkgreen}{\footnotesize(-3.45)} & 48.15 \textcolor{mydarkgreen}{\footnotesize(-1.10)} \\
\bottomrule[2pt]
\end{tabular}
}
\end{table}

\begin{table}[!t]
\centering
\caption{\textbf{Ablation study on loss components.} Results of Avg. mIoU (\%) are reported. $\mathcal{L}_{con}$: supervised contrastive loss, $\mathcal{L}_{chc}$: hierarchical consistency loss, $\mathcal{L}_{bis}$: bi-branch supervision loss.}
\label{tab:loss_ablation_study}
\vspace{-0.2cm}
\resizebox{\linewidth}{!}{ 
\begin{tabular}{cccccc}
\toprule[2pt]
\multicolumn{3}{c}{\textbf{Loss component}} & \multicolumn{3}{c}{\textbf{Dataset}} \\
\cmidrule(r){1-3} \cmidrule(r){4-6} 
\multirow{1}{*}{$\mathcal{L}_{con}$} & \multirow{1}{*}{$\mathcal{L}_{chc}$} & \multirow{1}{*}{$\mathcal{L}_{bis}$} & \multirow{1}{*}{\textbf{Campus3D}} & \multirow{1}{*}{\textbf{S3DIS-H}} & \multirow{1}{*}{\textbf{SensatUrban-H}} \\
\midrule
\rowcolor{gray!10} \ding{51} & \ding{51} & \ding{51} & 63.28 & 66.43 & 49.25 \\
 \ding{55} & \ding{51} & \ding{51} & 59.15 \textcolor{mydarkgreen}{\footnotesize(-4.13)} & 61.49 \textcolor{mydarkgreen}{\footnotesize(-4.94)} & 48.75 \textcolor{mydarkgreen}{\footnotesize(-0.50)} \\
 \ding{51} & \ding{55} & \ding{51} & 61.08 \textcolor{mydarkgreen}{\footnotesize(-2.20)} & 64.09 \textcolor{mydarkgreen}{\footnotesize(-2.34)} & 47.50 \textcolor{mydarkgreen}{\footnotesize(-1.75)} \\
 \ding{51} & \ding{51} & \ding{55} & 61.95 \textcolor{mydarkgreen}{\footnotesize(-1.33)} & 62.98 \textcolor{mydarkgreen}{\footnotesize(-3.45)} & 48.15 \textcolor{mydarkgreen}{\footnotesize(-1.10)} \\
\bottomrule[2pt]
\end{tabular}
}
\vspace{-0.2cm}
\end{table}

In this part, we conduct ablation studies to validate the effectiveness of key components in our framework.
Tables~\ref{ta} and \ref{tab:loss_ablation_study} report the results on three datasets using PointNet++.

\vspace{+0.05cm}
\noindent\textbf{Ablation study on model structure.}
From the perspective of model structure, the core designs in our framework include the late-decoupled framework (LDF), coarse-to-fine guidance (CFG), and auxiliary discrimination branch (ADB).
Table~\ref{ta} shows their ablation experiments.
First, removing LDF results in the most severe performance drop, e.g., the Avg.mIoU significantly decreases by $4.06\%$ on Campus3D and $4.26\%$ on S3DIS-H, respectively.
This demonstrates the crucial effect of our proposed late-decoupled framework in mitigating multi-hierarchy conflicts.
Second, disabling CFG also leads to a noticeable $3.32\%$ performance drop on Campus3D, validating the efficacy of using coarser hierarchies to guide the fine hierarchies.
Third, the removal of ADB also causes performance degradation across all datasets (e.g., $3.45\%$ drop on S3DIS-H), indicating our auxiliary discrimination branch is helpful for the fine-grained 3DHS tasks with class imbalance.

\vspace{+0.05cm}
\noindent\textbf{Ablation study on loss components.} 
From the perspective of optimization loss, the key components in our framework include the supervised contrastive loss ($\mathcal{L}_{con}$), hierarchical consistency loss ($\mathcal{L}_{chc}$), and bi-branch supervision loss ($\mathcal{L}_{bis}$).
Table~\ref{tab:loss_ablation_study} shows their ablation experiments.
Specifically, the contribution of $\mathcal{L}_{con}$ was particularly significant, e.g., its removal incurs a substantial mIoU drop of $4.94\%$ on the S3DIS-H dataset.
The removal of $\mathcal{L}_{bis}$ also causes a clear performance drop across all three datasets, i.e., a $1.33\%$ decrease on Campus3D, $3.45\%$ on S3DIS-H, and $1.10\%$ on SensatUrban-H.
Similarly, ablating $\mathcal{L}_{chc}$ also results in performance degeneration, e.g., $2.34\%$ mIoU drop on S3DIS-H.
These results demonstrate the necessary role of each component in our proposed Ld-3DHS framework.

\begin{figure}[t]
    \centering
    \begin{subfigure}[b]{0.49\columnwidth}
        \centering
        \includegraphics[width=\linewidth]{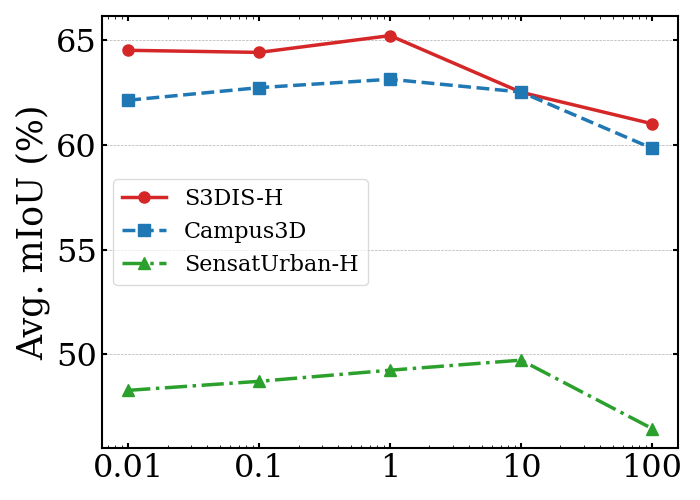}
        \caption{$\alpha$}
        \label{fig:fusion_weight}
    \end{subfigure}
    \hfill 
    \begin{subfigure}[b]{0.49\columnwidth}
        \centering
        \includegraphics[width=\linewidth]{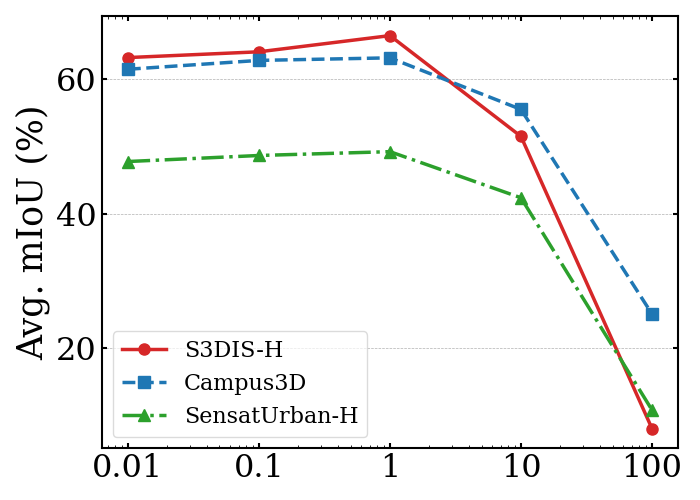}
        \caption{$\lambda$}
        \label{fig:loss_weight}
    \end{subfigure}
    \vspace{-0.25cm}
    \caption{Hyper-parameter analysis for (a) coarse-to-fine guidance weight $\alpha$ and (b) auxiliary discrimination branch loss weight $\lambda$.}
    \label{fig:parameter_analysis_horizontal}
    \vspace{-0.25cm}
\end{figure}

\vspace{+0.05cm}
\noindent\textbf{Hyper-parameter analysis.} Our method includes two hyper-parameters: the weight $\alpha$ in coarse-to-fine guidance in Eq.~(\ref{eq:feature_fusion}) and the weight $\lambda$ of auxiliary discrimination loss in Eq.~(\ref{eq:total_loss}).
The parameter analysis is shown in Figure~\ref{fig:parameter_analysis_horizontal}.
For $\alpha$, the performance peaks in a moderate range, i.e., $0.1<\alpha<10.0$.
For $\lambda$, we find it is insensitive at the range of $[0.1, 1.0]$.
For generality, we set $\alpha=1.0$ and $\lambda=1.0$ for comparison experiments on all datasets and backbones.


\vspace{-0.1cm}
\section{Conclusion}
\label{sec:conclusion}

In this paper, we focus on addressing two persistent but overlooked challenges for 3D hierarchical semantic segmentation: the multi-hierarchy conflict between learning multiple labels and a parameter-sharing model, and the class imbalance issue hindering performance on minority classes.
To this end, we propose {Ld-3DHS}, a novel late-decoupled framework with semantic prototype discrimination based bi-branch supervision.
The late-decoupled architecture can mitigate the multi-hierarchy conflicts and the auxiliary discrimination branch can improve the segmentation ability for point clouds with class imbalance.
Extensive experiments on multiple datasets and backbones demonstrate that our approach achieves state-of-the-art 3DHS performance, whose core components can also be used as a plug-and-play enhancement to improve previous 3DHS methods.

\vspace{+0.05cm}
\noindent\textbf{Limitation and future work.}
Although our {Ld-3DHS} is proven to be effective, the non-shared decoders and the extra auxiliary discrimination branch will increase the model size.
Therefore, future work can explore lightweight models and open-vocabulary strategies to further improve the efficiency and performance on extremely rare classes.



{
    \small
    \bibliographystyle{ieeenat_fullname}
    \bibliography{main}
}

\newpage

\section*{Appendix}

\subsection*{A. Implementation Details}

\subsubsection*{A.1. Key Hyperparameters}

Table~\ref{tab:hyperparameters} summarizes the key hyperparameters used in our Ld-3DHS framework.

\begin{table}[h]
\centering
\caption{Key hyperparameters in Ld-3DHS framework.}
\label{tab:hyperparameters}
\begin{tabular}{lcc}
\toprule
\textbf{Hyperparameter} & \textbf{Symbol} & \textbf{Value} \\
\midrule
EMA momentum & $\beta$ & 0.999 \\
Auxiliary loss weight & $\lambda$ & 1.0 \\
Guidance weight & $\alpha$ & 1.0 \\
Contrastive temperature & $\tau$ & 0.07 \\
Learning rate (initial) & $\eta$ & 0.01 \\
Learning rate (final) & - & 1.0e-5 \\
Weight decay & - & 1.0e-4 \\
Batch size (S3DIS) & $|B|$ & 12 \\
Batch size (Campus, Urban) & $|B|$ & 64 \\
Label smoothing factor & - & 0.2 \\
Training epochs & - & 100 \\
\bottomrule
\end{tabular}
\end{table}

\subsubsection*{A.2. Network Architecture}

This section details the network architecture of our Ld-3DHS model. The structural specifications for the main and auxiliary branches are presented in Table~\ref{tab:network_architecture}.







\begin{table}[t]
\centering
\caption{Network Architecture of Ld-3DHS. $K^{(h)}$ denotes the number of classes in hierarchy $h$.}
\label{tab:network_architecture}
\small 
\setlength{\tabcolsep}{4pt} 
\renewcommand{\arraystretch}{1.1} 
\begin{tabularx}{\columnwidth}{@{}lX@{}} 
\toprule
\textbf{Component} & \textbf{Specification} \\
\midrule
\multicolumn{2}{@{}l}{\textbf{Main Branch (Late-decoupled 3DHS)}} \\
\midrule
Encoder       & \textbf{PointNet++} (4 SA layers). \newline
              \textit{MLP Dims:} \texttt{[32,64]} $\rightarrow$ \texttt{[64,128]} $\rightarrow$ \texttt{[128,256]} $\rightarrow$ \texttt{[256,512]}. \newline
              \textit{Sampling:} 32 samples/layer (FPS), radius $r=0.1$. \\
\cmidrule(r){1-2}
Decoder       & \textbf{Feature Propagation}. \newline
              \textit{MLP Dims:} \texttt{[128,128,128]}, \texttt{[256,128]}, \texttt{[256,256]}, \texttt{[256,256]}. \\
\cmidrule(r){1-2}
Classifier    & \textbf{MLP} with Dropout (0.5). \textit{Dims:} \texttt{[128, 128, $K^{(h)}$]}. \\
\midrule
\multicolumn{2}{@{}l}{\textbf{Auxiliary Discriminative Branch}} \\
\midrule
Encoder       & \textbf{PointNet++} instance. \newline
              \textit{MLP Dims:} Same as the main branch encoder. \newline
              \textit{Sampling:} 16 samples/layer (vs. 32) for efficiency. \newline
              \textit{Output:} 512-dim features for contrastive learning. \\
\bottomrule
\end{tabularx}
\end{table}

\subsubsection*{A.3. Dataset Hierarchy Structures} 

The hierarchical class structures, detailing the mapping from coarse to fine-grained categories for each dataset, are provided in Table~\ref{tab:hierarchy_details}.

\subsubsection*{A.4. Contrastive Learning and Prototype Details}

\textbf{Contrastive Loss:} Temperature $\tau = 0.07$. Features are L2-normalized. Negative samples: all features from other classes in the same batch.

\noindent\textbf{Prototype Update:} The prototypes for each class are updated using Exponential Moving Average (EMA) with a high momentum coefficient $\beta = 0.999$, starting from the first epoch (no warm-up). This ensures a stable and smooth evolution of the class representations over time. The update rule for a prototype $P^{(h,c)}$ (representing class $c$ in hierarchy $h$) is defined as:
\begin{equation} \label{eq:ema_update}
P^{(h,c)} \gets \beta \cdot P^{(h,c)} + (1 - \beta) \cdot p^{(h,c)}
\end{equation}
where $p^{(h,c)}$ is the new prototype computed as the mean feature of all points belonging to class $c$ in the current mini-batch. Prototypes are computed using ground truth labels and are L2-normalized after each update step.

\subsubsection*{A.5. Gini-based Hierarchical Imbalance Activation}

To adaptively manage class imbalance within each hierarchy, we employ a dynamic activation strategy based on the Gini coefficient. This strategy assesses if a hierarchy's class point distribution is severely imbalanced to determine whether to activate an imbalance handling mechanism (e.g., the Auxiliary Discrimination Branch $\mathcal{L}_{\text{aux}}$).

For hierarchy $h$, we quantify the imbalance by calculating the Gini coefficient $G_h$ from its class frequencies $f_{h,i}$:
$$
G_h = \frac{\sum_{i=1}^{C_h} \sum_{j=1}^{C_h} |f_{h,i} - f_{h,j}|}{2C_h}
$$
where $C_h$ is the total number of classes at hierarchy $h$, and $f_{h,i}$ is the point frequency of class $i$. The value of $G_h$ ranges from 0 (perfect balance) to nearly 1 (maximum imbalance).

We set an activation threshold $\tau$ (e.g., 0.6). The imbalance handling module is activated only when $G_h \ge \tau$:
$$
\text{Activate}_{\text{Imbalance-h}} = (G_h \ge \tau)
$$
This hierarchy-specific dynamic intervention allows the model to efficiently and selectively address severe class imbalance issues.

\subsection*{B. Detailed Algorithms}

In this section, we provide a detailed breakdown of the algorithms used in our proposed Ld-3DHS framework.

\subsubsection*{B.1. Overall Training Framework}

The Algorithm~\ref{alg:overall_simplified} outlines the overall training framework of our proposed Ld-3DHS. It demonstrates how the main branch (Ld-3DHS Branch) and the auxiliary branch (Auxiliary Discrimination Branch) are jointly optimized in each training iteration.

\begin{algorithm}[!h]
\caption{: Training Framework for Ld-3DHS}
\label{alg:overall_simplified}
\begin{algorithmic}[1]
\State \textbf{Input:} 3D point cloud dataset $\mathcal{X}$, hierarchical labels $\{Y^{(h)}\}_{h=1}^H$
\State \textbf{Parameters:} Learning rate $\eta$, batch size $|B|$, trade-off $\lambda$, EMA $\beta$
\State Initialize network $f_\theta$ and prototypes $\{P_{\text{main}}^{(h)}\}, \{P_{\text{aux}}^{(h)}\}$

\While{not reaching the maximal epochs}
    \State Sample a mini-batch $B \subset \mathcal{X}$
    \State Extract shared features using the encoder of $f_\theta$
    
    \State \textit{// 1. Process Main Branch}
    \For{$h = 1 \to H$}
        \State Compute main branch logits $L^{(h)}$, applying coarse-to-fine guidance from $L^{(h-1)}$ if $h>1$.
    \EndFor
    \State Compute main branch loss $\mathcal{L}_{\text{late-3DHS}}$ using cross-entropy ($\mathcal{L}_{\text{ces}}$) and consistency ($\mathcal{L}_{\text{chc}}$) objectives.
    
    \State \textit{// 2. Process Auxiliary Branch}
    \For{each hierarchy $h$}
        \State Extract class-specific auxiliary features $\{F_{\text{aux}}^{(h,c)}\}$.
    \EndFor
    \State \textbf{with no gradient:}
    \State \hspace{\algorithmicindent} Update all main and auxiliary prototypes via EMA.
    \State Compute auxiliary branch loss $\mathcal{L}_{\text{aux}}$ by summing the Contrastive Loss ($\mathcal{L}_{\text{con}}$) and Bi-Branch Supervision Loss ($\mathcal{L}_{\text{bis}}$) over all hierarchies.

    \State \textit{// 3. Joint Optimization and Prototype Updates}
    \State Compute the final total loss $\mathcal{L}_{\text{total}} \gets \mathcal{L}_{\text{late-3DHS}} + \lambda \mathcal{L}_{\text{aux}}$.
    \State Update network parameters $\theta$ via backpropagation on $\mathcal{L}_{\text{total}}$.
\EndWhile
\State \textbf{Output:} The trained model parameters $\theta$
\end{algorithmic}
\end{algorithm}

\newpage

\begin{table}[!h]
\centering
\caption{Detailed Hierarchy Structures. Mappings for Campus3D are based on the full 5-level hierarchy structure.}
\label{tab:hierarchy_details}
\small
\setlength{\tabcolsep}{4pt} 
\renewcommand{\arraystretch}{1.2} 
\begin{tabularx}{\columnwidth}{@{}l|X@{}} 
\toprule
\textbf{Coarse Category (Parent)} & \textbf{Mapped Fine-Grained Categories (Children)} \\
\midrule[1pt]
\multicolumn{2}{@{}l}{\textbf{S3DIS-H}} \\ 
\addlinespace[2pt]
(A) Static Elements & wall, floor, ceiling \\
(B) Openings & window, door \\
(C) Furniture & table, chair, sofa, bookcase \\
(D) Miscellaneous & beam, column, board, clutter \\
\midrule[1pt]
\multicolumn{2}{@{}l}{\textbf{SensatUrban-H}} \\
\addlinespace[2pt]
Core Traffic Infrastructure & Road, Footpath, Parking, Bridge, Rail \\
Natural \& Ground & Vegetation, Ground, Water \\
Traffic Elements & Car, Bike, Unclassified \\
Urban Amenities & Building, Street Furniture, Wall \\
\midrule[1pt]
\multicolumn{2}{@{}l}{\textbf{Campus3D} (Based on L1, L3, L5)} \\
\addlinespace[2pt]
\multicolumn{2}{@{}l}{\textit{Level 1 $\rightarrow$ Level 3 Mapping:}} \\
\addlinespace[2pt]
L1: ground & L3: natural, play\_field, path\&\text{stair}, driving\_road, man\_made \\
L1: construction & L3: construction \\
L1: unclassified & L3: unclassified \\
\cmidrule(l){1-2}
\addlinespace[2pt]
\multicolumn{2}{@{}l}{\textit{Level 3 $\rightarrow$ Level 5 Mapping:}} \\
\addlinespace[2pt]
L3: natural & L5: natural \\
L3: play\_field & L5: play\_field, sheltered, unsheltered, bus\_stop \\
L3: path\&\text{stair} & L5: path\&\text{stair} \\
L3: driving\_road & L5: car, bus, not\_vehicle \\ 
L3: man\_made & L5: man\_made \\
L3: construction & L5: roof, wall, link, artificial landscape, lamp, others \\ 
L3: unclassified & L5: miscellaneous \\
\bottomrule[1pt]
\end{tabularx}
\end{table}

\end{document}